\documentclass[conference]{IEEEtran}
\IEEEoverridecommandlockouts
\usepackage{cite}
\usepackage[numbers]{natbib}
\usepackage{amsmath,amssymb,amsfonts}
\usepackage{algorithmic}
\usepackage{graphicx}
\usepackage{tabularx,booktabs}
\usepackage{multirow}
\usepackage{makecell}
\usepackage{booktabs}
\usepackage{textcomp}
\usepackage{siunitx}
\usepackage{subcaption}
\usepackage{todonotes}
\usepackage{lipsum}
\usepackage{hyperref}
\usepackage{xcolor}
\def\BibTeX{{\rm B\kern-.05em{\sc i\kern-.025em b}\kern-.08em
    T\kern-.1667em\lower.7ex\hbox{E}\kern-.125emX}}

\newcolumntype{Y}{>{\centering\arraybackslash}X}
\newcommand{\N}{\hphantom{0}}

\makeatletter
\newcommand{\linebreakand}{%
  \end{@IEEEauthorhalign}
  \hfill\mbox{}\par
  \mbox{}\hfill\begin{@IEEEauthorhalign}
}
\makeatother

\begin{document}

\title{Zero-Shot Text Matching for Automated Auditing using Sentence Transformers\\
}

\author{
\IEEEauthorblockN{1\textsuperscript{st} David Biesner}
\IEEEauthorblockA{\textit{Fraunhofer IAIS and University of Bonn} \\
Sankt Augustin and Bonn, Germany \\
david.biesner@iais.fraunhofer.de}
\and
\IEEEauthorblockN{2\textsuperscript{nd} Maren Pielka}
\IEEEauthorblockA{\textit{Fraunhofer IAIS} \\
Sankt Augustin, Germany}
\and
\IEEEauthorblockN{3\textsuperscript{rd} Rajkumar Ramamurthy}
\IEEEauthorblockA{\textit{Fraunhofer IAIS and University of Bonn} \\
Sankt Augustin and Bonn, Germany}
\and
\IEEEauthorblockN{4\textsuperscript{th} Tim Dilmaghani}
\IEEEauthorblockA{\textit{PWC GmbH WPG} \\
Frankfurt a. M., Germany}
\and
\IEEEauthorblockN{5\textsuperscript{th} Bernd Kliem}
\IEEEauthorblockA{\textit{PWC GmbH WPG} \\
Frankfurt a. M., Germany}
\and
\IEEEauthorblockN{6\textsuperscript{th} Rüdiger Loitz}
\IEEEauthorblockA{\textit{PWC GmbH WPG} \\
Frankfurt a. M., Germany}
\and
\IEEEauthorblockN{7\textsuperscript{th} Rafet Sifa}
\IEEEauthorblockA{\textit{Fraunhofer IAIS} \\
Sankt Augustin, Germany}
} 

\maketitle

\begin{abstract}
Natural language processing methods have several applications in automated auditing, including document or passage classification, information retrieval, and question answering. However, training such models requires a large amount of annotated data which is scarce in industrial settings. At the same time, techniques like zero-shot and unsupervised learning allow for application of models pre-trained using general domain data to unseen domains. 

In this work, we study the efficiency of unsupervised text matching using Sentence-Bert, a transformer-based model, by applying it to the semantic similarity of financial passages. Experimental results show that this model is robust to documents from in- and out-of-domain data.\footnote{To be published in proceedings of IEEE International Conference on Machine Learning Applications IEEE ICMLA 2022.}
\end{abstract}


\section{Introduction}

The auditing process of financial reports involves proofreading and correcting yearly financial statements by trained auditors.
The auditors present an opinion on whether the report issued by the audited company is prepared according to all legal requirements of the applicable financial reporting framework such as International
Financial Reports Standards (IFRS) and Handelsgesetzbuch (HGB). This process requires expert knowledge, judgment and experience by the auditors.

However, a large part of the described process proves itself recurrent, manual and time-consuming. For example, consider the evaluation of a financial report to the guidelines given by a regulatory framework.
The guidelines are generally presented as a series of checklist requirements that must be addressed in the report.
For each such checklist item, the designated auditor has to first find the relevant section in the document and then use their expert judgement to answer the compliance question raised by the checklist item. Finding pertinent sections of a large document can be time-consuming since many items do not lend themselves to a simple lookup function and relevant passages can be spread throughout the text.
For regulatory frameworks with a couple hundred to over a thousand requirements and report lengths of up to several hundred pages,
the simple process of finding relevant text passages can take up a large portion of an auditor's time.

One possible approach to addressing this issue is using machine learning algorithms for text recommendation. 
That is, given a specific requirement text, the goal of the prediction model is to recommend one or more relevant text passages, ideally as a list ranked by relevance to the desired requirement. 
Previous work \cite{ali,ramamurthy2021alibert} has formulated this problem as a text classification problem, in which the model determines the relevance of a text passage from a financial report to a specific requirement in the checklist. First, the text passage is encoded to a latent representation (using TF-IDF \cite{ali} or transformer-based language models \cite{ramamurthy2021alibert}). 
Then this representation is used by a classification model (a multi-layer perceptron or a recurrent neural network) to assign relevance scores to each requirement from the checklist.
Once relevance scores for all pairs of passage-requirement are computed, the results can be ranked by relevance for each requirement.

\begin{figure}
    \centering
    \includegraphics[width=0.9\columnwidth]{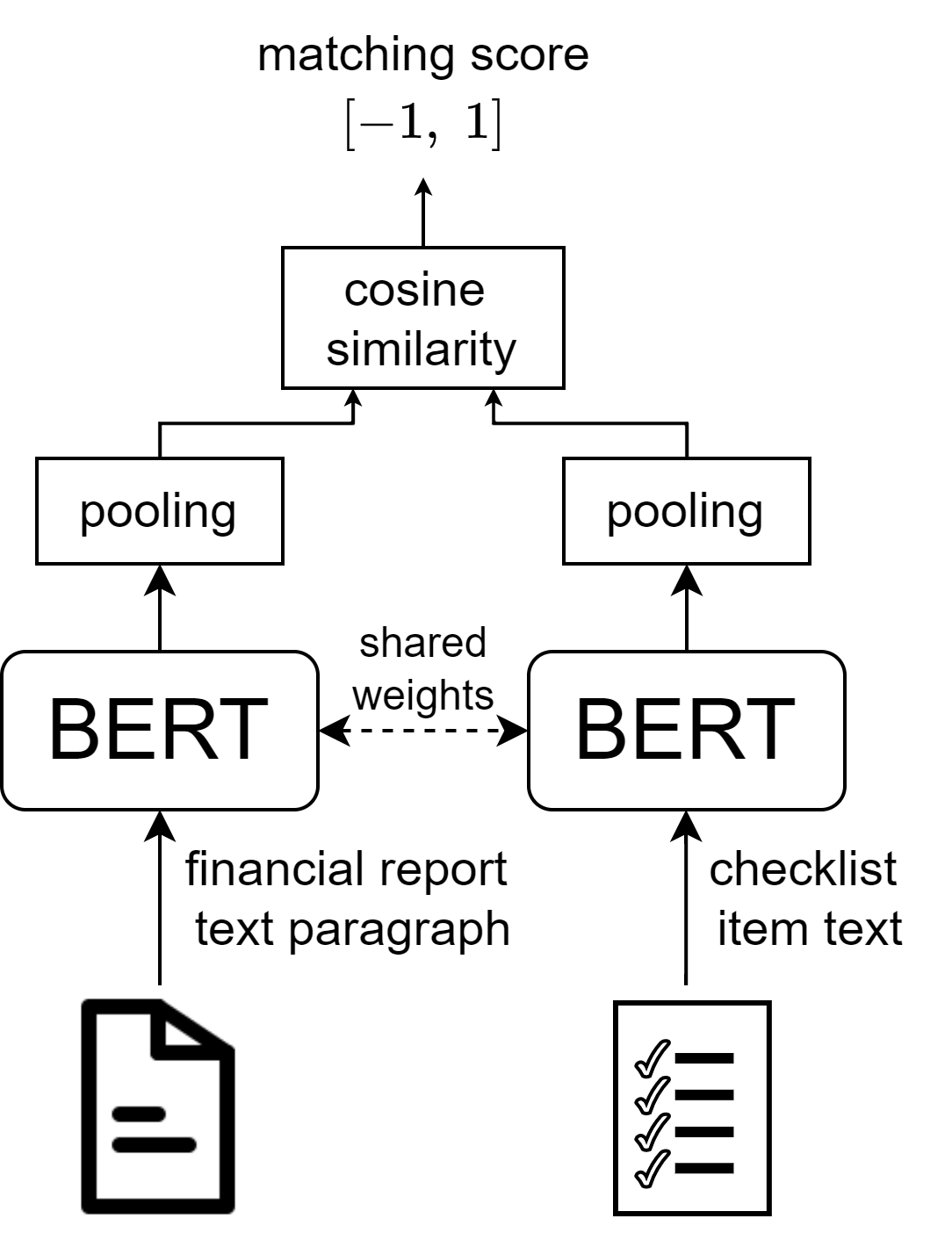}
    \caption{Overview of the SentenceBERT architecture for matching of financial report texts and checklist items. The model contains two BERT models with shared model weights which embed the financial report text and the description text of the checklist item. The output embeddings of BERT are mean-pooled and the cosine similarity score is calculated to evaluate whether both items match.}
    \label{fig:model}
\end{figure}

The detriment of this process is two-fold: on the one hand, training such a model requires annotated data, i.e., financial reports with text paragraphs annotated as relevant for specific requirements or a list of relevant text passages from various reports for each requirement.
Curating this data is time-consuming and therefore expensive.
On the other hand, these models deal with a fixed set of possible requirements from a checklist. Once a classification layer or a group of requirement classifiers is trained, adding new requirements to predict is only possible by retraining of the partial or the entire model.
However, legal frameworks change over time and it is therefore possible that a model trained on data from a particular year might not be usable in successive years.

Additionally, these models ignore a critical aspect of the auditing process:
requirements are not only target labels for the classification model,
they are written text and therefore contain information that might be useful for a model to understand the task better.
A human auditor does not memorize which text passages are relevant for which checklist entry number but rather processes the requirement text and tries to understand which text passages might match. Therefore, a model that considers text from a textual passage as well as the requirement to compute relevance scores is desired.

To alleviate these issues, we propose a recommendation approach that can be trained entirely unsupervised, without annotated data, and can recommend target requirements without ever seeing them during training.
Based on textual semantic similarity, our proposed model encodes both the requirement and paragraph text to latent representations and calculates the cosine similarity between the two vectors to get a relevance score.
Semantically close texts are assigned latent vectors in similar regions of latent space with high cosine similarity,
texts that are not related are assigned latent vectors with low similarity.
This method can be used entirely without domain-specific training.
A language model pre-trained on general language data will assign latent vectors with the latent space similarity property described above to any text and can be used to rank text paragraphs for relevance out of the box.
Further training of the model to better encode and match domain-specific texts can be done both in a supervised and unsupervised manner.

Such a model is then capable of matching texts to entirely new requirements it has never seen during training, simply by following the steps above and encoding the new requirement text.

In this work, we
\begin{itemize}
    \item further examine previous work in the field of automated auditing and work related to our model architecture, 
    \item describe the specific datasets we use for training and evaluation in our experiments and give details on the evaluation checklist,
          as well as details on the proposed model architecture and training steps,
    \item show how multiple training stages, both unsupervised and supervised, contribute to improvements in classification performance for labels seen during training and not seen during training,
    \item evaluate the effectiveness of two distinct types of unsupervised training methods for sentence encoding,
    \item demonstrate how the proposed text matching based algorithm is a capable recommendation system for both known and new requirements, and
    \item give an outlook into further research and integration and application of our methods.
\end{itemize}

\section{Related Work}
There has been some prior work on the topic of machine learning in the field of auditing financial reports. In \cite{ali}, the authors present a tool that can partially automate the auditing process by recommending to the user the most relevant text passages in a report for each legal requirement. In follow-up work, \cite{ramamurthy2021alibert} builds upon those findings, improving the model by adding fine-tuned BERT \cite{devlin2018bert} embeddings and an end-to-end training setup.

Another direction of research in this regard is the extraction of key performance indicators and their corresponding numerical values \cite{kpibert} from the text, with the goal to cross-check the consistency of reported numbers throughout a document. A similar approach has been proposed by \cite{cao2018towards}, aiming to extract formulas from the text.

There is also the necessity for sensitive content, which is a highly relevant topic in the context of financial reports. To this end, \cite{biesner2020leveraging} presents a Named Entity Recognition-based anonymization approach for financial data. Further, there is also some prior work on sentiment analysis for financial risk prediction, e.g. by  \cite{tsai2017risk} and \cite{campbell2014information}.  

In our work, we build upon earlier work \cite{ali, ramamurthy2021alibert} and propose to use SentenceBERT \cite{reimers-2019-sentence-bert} architecture specialized for semantic similarity tasks.

\begin{table}[]
    \centering
    \begin{tabular}{ll}
        C\_1\_1 & \makecell[l]{Provide a description of each group of biological assets \\(narrative or quantified description).} \\
        E\_1\_7 & \makecell[l]{Disclose: (e) reconciliations of changes in insurance liabilities,\\ reinsurance assets and, related deferred acquisition costs, if any.} \\
        F\_1\_9 & \makecell[l]{For defined benefit plans, disclose: (a) significant actuarial \\assumptions made; (b) date of the most recent actuarial valuation.} \\
    \end{tabular}
    \captionof{figure}{Examples for checklist items in the IFRS regulatory framework. The framework contains a total of 1305 individual items.}
    \label{tab:ifrs}
\end{table}

\begin{table}[]
\begin{tabular}{llrrr}
\toprule
Dataset                   & Split       & Paragraphs & Words     & Requirements  \\
\midrule
BANZ                      & -           & \num{70203}      & \num{1144226}   & -            \\
SEC               & -           & \num{70977}      & \num{19274397}  & -            \\ \\
\multirow{3}{*}{IFRS\_DE} & train       & \num{20522}    & \num{1234917} & \num{758}        \\
                          & val         & \num{2281}     & \num{160180}  & \num{761}        \\
                          & test seen   & \num{2818}     & \num{175747}  & \num{758}        \\
                          & test unseen & \num{177}      & \num{12593}   & \num{8}          \\ \\
\multirow{3}{*}{IFRS\_EN} & train       & \num{12486}    & \num{813854}  & \num{669}        \\
                          & val         & \num{1744}     & \num{119562}  & \num{491}        \\
                          & test seen   & \num{1869}     & \num{140768}  & \num{669}       \\
                          & test unseen & \num{272}      & \num{21333}   & \num{7}          \\
\bottomrule
\end{tabular}
\caption{Statistics on the dataset applied in the training and evaluation process.
We consider two datasets of un-annotated financial text, in German (\emph{BANZ} from the Bundesanzeiger) and English (\emph{SEC} from the US Securities and Exchange Commision),
and two datasets of financial reports, annotated with corresponding IFRS requirements,
in German (\emph{IFRS\_DE}) and English (\emph{IFRS\_EN}).
The un-annotated sets are only used for unsupervised training and not split,
the annotated datasets are split into training, validation and test.
The \emph{test seen} split contains annotations from requirements also present in the training set,
the \emph{test unseen} split contains only additional requirements not present in the training set.}
\label{tab:data}
\end{table}

\section{Model Architectures}

Our proposed method for text matching is based on the SentenceBERT architecture \cite{reimers-2019-sentence-bert}. See Figure \ref{fig:model} for a diagram of the model architecture.
The architecture is an adjustment of the well-known BERT \cite{devlin2018bert} language modeling architecture and compatible with all pre-trained BERT models.
BERT is a transformer-based \cite{transformer} language model that is trained by reconstructing masked tokens of an input text from large datasets of un-annotated text.
The language model outputs an embedding for each token in the input text, which is used during training to predict the masked tokens, and can be used for various downstream tasks like token classification in named entity recognition.
The SentenceBERT model encodes both input texts via two BERT models with shared model weights.
To convert the token embeddings to sentence embeddings, we apply a mean-pooling layer.

To evaluate how well the two input texts match, the cosine similarity between the two embeddings is calculated,
which gives a score between $-1.0$ (low similarity) and $1.0$ (high similarity).

Note that while the model architecture is called SentenceBERT,
the model is not restricted to a single sentence as input but can embed any paragraph or text up to 512 subword tokens.

To predict and recommend checklist items for a given paragraph, 
the paragraph and all checklist item texts are encoded by the model.
The cosine similarity between the paragraph and each checklist item is calculated and sorted.
The model outputs the top-$k$ checklist items with the highest similarity score.
Analogously, one can recommend report text paragraphs for a specific checklist item by encoding all paragraphs in the report and calculating their cosine similarity with the encoded checklist item text.

\section{Data}
\label{sec:data}
In this section, we briefly describe the data we used in out experiments.
See Table \ref{tab:data} for an overview of all datasets and training, validation and test splits.

For unsupervised training,
we provide two datasets of German and English language.
The German dataset consists of financial reports from BundesAnzeiger\footnote{\url{https://www.bundesanzeiger.de/}} (\emph{BANZ}).
The English dataset consists of financial reports from the US Securities and Exchange Commision\footnote{\url{https://www.sec.gov/dera/data/financial-statement-and-notes-data-set.html}} (\emph{SEC}).
We only use these datasets for training, not evaluation and therefore do not split them.
See Table \ref{tab:data} for a reference on number of paragraphs and word counts.

For supervised training,
we annotate two datasets of financial reports in German and English language.
Those data sets were provided to us by an auditing firm, who collaborated with us for this project. 
We annotate each paragraph as, if applicable, corresponding to one or more individual checklist items from
the IFRS regulatory checklist with a total of \num{1305} items.
See Figure \ref{tab:ifrs} for sample checklist items with corresponding description text. 
We keep only annotated paragraphs in the dataset, 
see Table \ref{tab:data} for an overview of number of paragraphs and word count in each split.
We additionally describe the number of individual requirements that are part of some annotation
in the corresponding split.

For model testing, we create two test splits.
One test set stems from the same distribution as the training set. 
This test set contains no report texts that are also in the training set,
but the requirement annotations are for the same requirements as the annotations in the training set.
That way, the model has already seen report texts that match the requirements in question.
We call these test sets \emph{test seen} (see Table \ref{tab:data}).

The other type of test set contains new report texts and requirement annotations that the model has not seen during training, i.e. no report text in the training dataset has any annotation of these requirements.
To predict reasonable text matches, the model must understand the requirement texts itself and match the fitting texts correctly.
A binary or multi-label classification model trained to predict the requirements in the training dataset cannot predict any of these requirements.

\section{Training and Transfer Learning}

The proposed method allows for various types of supervised and unsupervised learning.
We therefore train the model in multiple steps on various datasets and evaluate the performance after each training stage.

\subsection{Language Model Training}
\label{sec:lm_training}
The underlying BERT architecture of the SentenceBERT model itself is a language model,
meaning it is trained to reconstruct un-annotated text input and learn from datasets of raw text.
For details on the training of a BERT language model we refer to \cite{devlin2018bert}.

BERT is available pretrained on a variety of datasets.
We consider the model \emph{bert-base-multilingual-cased}\footnote{\url{https://huggingface.co/bert-base-multilingual-cased}},
which is trained on Wikipedia text from 104 languages.
The model should be able to process any type of input text in a reasonable manner to provide a baseline for further training.
Results for models based on this language model are shown in Table \ref{tab:results_general}.

Additionally, we finetune the \emph{bert-base-multilingual-cased} language model on a dataset of financial language data (\emph{BANZ} and \emph{SEC}, see Table \ref{tab:data}).
We theorize that a language model trained on domain specific data will improve matching performance after further training steps.
We show results for model based on this language model in Table \ref{tab:results_financial}.

\subsection{Unsupervised Training}
\label{sec:unsup_training}

The second training step consists of training the sentence embeddings in an unsupervised manner.
During this training stage, we do not differentiate between report texts and requirements and consider both as individual input texts.
We consider two training mechanisms: Simple Contrastive Learning of Sentence Embeddings (\emph{SimCSE} \cite{simcse}) and Tranformer-based Denoising AutoEncoder (\emph{TSDAE} \cite{tsdae}).

During the SimCSE training, we aim to improve the sentence embeddings by enforcing that each embedding occupies
its own region in latent space, meaning that embeddings for two different input sentences are far apart in latent space.
This has shown to significantly improve the performance of sentence embedding models in downstream tasks \cite{simcse}.
For training the model via SimCSE, we embed all sentences and requirements in a batch of documents and calculate the
cosine similarity between each pair of texts.
We add some noise using dropout in the encoding process, such that two embeddings of the same text are not exactly alike.
We train the model to maximize the cosine similarity between two embeddings of the same text and minimize the cosine similarity between embeddings from different texts.
For details on this training we refer to \cite{simcse}.

The second unsupervised training approach, TSDAE, is similar to the language modeling training described in Section \ref{sec:lm_training}.
For this, the model is adjusted into an encoder-decoder-structure.
The encoder is given by the SentenceBERT model trained in the previous training steps.
The decoder is given by a new SentenceBERT model, initialized either from the general language (\emph{bert-base-multilingual-cased}) or the financial data model.
During training, noise is added to each input text by masking a fraction of all tokens (we mask 60\% in our experiments),
encoding the noisy text and reconstructing the embedding using the decoder module.
After training, the decoder module is discarded and the encoder is used for prediction or further finetuning.
For details on this training we refer to \cite{tsdae}.

We train the model using both methods on two un-annotated datasets of German (\emph{BANZ} in Table \ref{tab:data}) and English (\emph{SEC}) language until convergence,
and select the model with the best matching score on the validation set for further training or evaluation.

\subsection{Supervised Training}

During the last training step, we aim to utilize the annotated datasets,
i.e. the datasets of financial reports in German and English with
report paragraphs annotated as matching a certain requirement text.
The training procedure is based on the same SimCSE method described in Secion \ref{sec:unsup_training}.
In this case, we do not want to only enforce two noisy encodings of the same text to be close in latent space,
but the encodings of matching report text and requirement pairs to be close, i.e. have a large cosine similarity.
In the same manner, we enforce non-matching text and requirement to have encodings with small cosine similarity.
For this, we encode all report texts and requirements from a batch of annotated data and calculate the cosine similarity between each report text and each requirement text.
Unlike the unsupervised training, we do not calculate the cosine similarity between two encodings of reports texts or two encodings of requirement texts.
We train the model to maximize the cosine similarity of matching text and requirement pairs and minimize the cosine similarity of all other pairs.
For further details on the supervised training we again refer to \cite{simcse}.

We train the model on German, English or German and English annotated reports (\emph{DE}, \emph{EN} and \emph{DE+EN} in Table \ref{tab:data}, respectively)
until convergence and select the model with the best matching score on the validation set for evaluation.

\section{Evaluation and Results}

\begin{table*}[]
\centering
    \begin{subtable}{\linewidth}
    \centering
        
\begin{tabularx}{0.7\textwidth}{ll *{6}{Y}}
\toprule
        \multicolumn{8}{c}{Base Model: General Language BERT}                                                                                                                                                              \\ 
        \midrule
                             &                     & \multicolumn{2}{c}{DE}                                & \multicolumn{2}{c}{EN}                                & \multicolumn{2}{c}{DE+EN}                             \\ 
                             \cmidrule(lr){3-4} \cmidrule(l){5-6}  \cmidrule(l){7-8} 
        Unsupervised Method  & Supervised Training & \multicolumn{1}{c}{Seen} & \multicolumn{1}{c}{Unseen} & \multicolumn{1}{c}{Seen} & \multicolumn{1}{c}{Unseen} & \multicolumn{1}{c}{Seen} & \multicolumn{1}{c}{Unseen} \\ 
        \midrule
        \multicolumn{2}{c}{Language Modeling Only} &  \N2.83  &  \N0.00  &  \N5.83  &  \N2.94  &  \N4.03  &  \N1.78  \\ 
                             &                     &                          &                            &                          &                            &                          &                            \\
        TSDAE                & -                   &  10.78   &  \N2.25  &  17.87  &  \N5.14  &  13.61  &  \N4.00  \\ 
        SIMCSE               & -                   &  \N4.86  &  \N1.69  &  \N8.29  &  \N5.14  &  \N6.22  &  \N3.78  \\ 
                             &                     &                          &                            &                          &                            &                          &                            \\
        TSDAE                & DE                  &  \textbf{91.55}  &  16.38  &  55.48  &  \textbf{66.17}  &  77.17  &  46.54  \\
        TSDAE                & EN                  &  52.23  &  34.46  &  89.08  &  20.58  &  66.92  &  26.05  \\
        TSDAE                & DE+EN               &  90.24  &  \textbf{39.54}  &  \textbf{92.45}  &  52.94  &  \textbf{91.12}  &  \textbf{47.66}  \\
        SIMCSE               & DE                  &  89.63  &  17.51  &  44.72  &  51.47  &  71.73  &  38.08  \\
        SIMCSE               & EN                  &  32.82  &  \N7.34  &  86.51  &  27.57  &  54.23  &  19.59  \\
        SIMCSE               & DE+EN               &  88.21  &  30.50  &  90.79  &  \textbf{52.94}  &  89.24  &  44.09  \\

        
        
        
        \bottomrule
\end{tabularx}
    \caption{Results for the BERT model pretrained on general language data}
    \label{tab:results_general}
    \end{subtable}
\\
\vspace{0.4cm}
    \begin{subtable}{\linewidth}
    \centering

\begin{tabularx}{0.7\textwidth}{ll *{6}{Y}}
        \toprule
        \multicolumn{8}{c}{Base Model: Financial Language BERT}                                                                                                                                                              \\ 
        \midrule
                             &                     & \multicolumn{2}{c}{DE}                                & \multicolumn{2}{c}{EN}                                & \multicolumn{2}{c}{DE+EN}                             \\ 
                             \cmidrule(lr){3-4} \cmidrule(l){5-6}  \cmidrule(l){7-8} 
        Unsupervised Method  & Supervised Training & \multicolumn{1}{c}{Seen} & \multicolumn{1}{c}{Unseen} & \multicolumn{1}{c}{Seen} & \multicolumn{1}{c}{Unseen} & \multicolumn{1}{c}{Seen} & \multicolumn{1}{c}{Unseen} \\ 
        \midrule
        \multicolumn{2}{c}{Language Modeling Only} &  \N0.69  &  \N0.00  &  \N0.61  &  \N0.10  &  \N0.65  &  \N0.66  \\
                             &                     &                          &                            &                          &                            &                          &                            \\
        TSDAE                & -                   &  12.31  &  \N0.12  &  17.81  &  \N0.51  &  14.50  &  \N0.78  \\
        SIMCSE               & -                   &  13.41  &  18.07  &  16.42  &  \N0.41  &  14.61  &  \N0.79  \\
                             &                     &                          &                            &                          &                            &                          &                            \\
        TSDAE                & DE                  &  \textbf{91.55}  &  33.89  &  52.43  &  \textbf{56.25}  &  75.95  &  \textbf{47.43}  \\
        TSDAE                & EN                  &  56.06  &  44.06  &  86.14  &  26.83  &  68.06  &  33.63  \\
        TSDAE                & DE+EN               &  88.57  &  46.32  &  \textbf{92.83}  &  45.58  &  \textbf{90.27}  &  45.87  \\
        SIMCSE               & DE                  &  88.39  &  34.46  &  50.82  &  45.22  &  73.41  &  40.97  \\
        SIMCSE               & EN                  &  51.59  &  30.50  &  87.74  &  30.14  &  66.01  &  30.28  \\
        SIMCSE               & DE+EN               &  83.67  &  \textbf{47.45}  &  84.43  &  46.69  &  83.97  &  46.99  \\

    
    

        \bottomrule
\end{tabularx}

    \caption{Results for the BERT model pretrained on financial documents}
    \label{tab:results_financial}
    \end{subtable}
\caption{Evaluation of all training runs on the hold-out test dataset.
We compare the models trained purely as language models, 
trained unsupervised using the TSDAE and SIMCSE method,
and further trained supervised on German, English or German and English data.
We evaluate the one-shot recall for the top-5 recommendations on the hold-out test sets.
We differentiate between German, English or German and English test data and seen and unseen requirements.
Note that for example for a model trained supervised only on German data, all English data is unseen.}
\label{tab:results}
\end{table*}

We evaluate all trained models on two types of hold-out test sets,
the test set with \emph{seen} requirements and the test set with \emph{unseen} requirements,
as described in Section \ref{sec:data}.

We evaluate each type of test set in German, English and combined German and English language (\emph{DE}, \emph{EN} and \emph{DE+EN} in Table \ref{tab:data}, respectively). 

We collect all evaluation results in Tables \ref{tab:results_general} and \ref{tab:results_financial}.
For each training step, we compute the one-shot recall on the top-5 recommendations given by the model,
which is a custom performance measure that has been designed to reflect user experience.
For each annotated report text, we let the model predict the matching score for all requirements
and select the 5 highest rated requirements as prediction.
We assign a score of $1.0$ for this sample if a correct requirement is part of this prediction and average the score over the dataset.
A 90\% one-shot  recall signifies that in 90\% of all cases, a correct requirement was part of the top-5 predictions.

Comparing both subtables \ref{tab:results_general} and \ref{tab:results_financial},
we see that pretraining the language model on financial data significantly improves classification for two of the three \emph{unseen} requirement datasets,
while the best metrics for \emph{seen} requirement dataset are very similar, increasing or decreasing by less than 1 percentage point depending on the dataset.
Also note that the same training procedures lead to similar results for both language models,
i.e. TSDAE unsupervised training and \emph{DE} supervised training performs best on the \emph{DE} \emph{seen} test set and TSDAE unsupervised training and \emph{DE+EN} supervised training best on both other \emph{seen} test sets.
Although this effect is less present when skipping the supervised training and not occuring when skipping both unsupervised and supervised training, we conclude a net benefit from pretraining the language model on domain specific data. 

For consecutive evaluations we only consider the model pretrained on financial language data in Table \ref{tab:results_financial}.

In order to evaluate the effect of both unsupervised training methods, SimCSE and TSDAE,
we first note that only unsupervised training results in poor performance on the test sets,
with a maximum of 17.81\% one-shot recall on a \emph{seen} test set.
We therefore concentrate on the effect of the unsupervised training method when continuing training in a supervised manner, see the bottom half of Table \ref{tab:results_financial}.

We see that the best TSDAE model outperforms the best SimCSE model by a significant margin on all \emph{seen} test sets (improvement of 3 to 5 percentage points) and two of the three \emph{unseen} test sets (improvement of 7 to 10 percentage points).
For the remaining \emph{unseen} test set the metric scores of the best models are very similar (46.32\% for TSDAE and 47.45\% for SimCSE).
As a result of this evaluation, we conclude that TSDAE is the more fitting unsupervised training method.

Finally, we consider the effect of supervised training data language on the test performance.
In this evaluation, we see no significant trend towards any specific data language setup.
While the model trained on German language only performs best on the German \emph{seen} test set,
the model trained on English language only does not outperform the other data setups in any test set.
But training on any English language data is clearly necessary for the \emph{seen} test sets in English or combined English and German.
Peculiarly, the model trained only on German data performs best on the English only \emph{unseen} test set,
with a significant margin against the next best model.

We conclude that, if available, training on data from both languages is recommended if the use case requires processing of texts in both languages.
However, more testing of this hypothesis is required.

In general, the evaluation shows that the proposed model architecture performs very well on the
task of predicting requirements available in training data (over 90\% one-shot recall on all \emph{seen} datasets).
Additionally, the methods produce reasonable results when faced with new requirements,
which a multilabel prediction model could not process at all.

In practice, such a model can be applied to replace a multilabel prediction model with little to no loss in performance.
When faced with a new or adjusted set of requirements, this model would not need to be retrained entirely but could be applied directly to the new task and improved by supervised training when training data is available.

\section{Conclusion and Outlook}

In this study, we analyzed the potential of sentence embedding methods for text matching based recommendation models in the domain of financial reports.
We trained a SentenceBERT model in a multi-stage training approach to embed sentences from financial reports and the text from corresponding regulatory checklist items and used the model to recommend 
labels that have been seen during training and labels that have never been seen during training.

We found that the proposed method performs competitively on \emph{seen} labels and reasonably well on \emph{unseen} labels.
This suggests these models to be useful in a setting were a subset of the labels or the entire label set might change between retrainings of a model.

Open points for future research include:
\begin{itemize}
    \item The application of the models in languages other than German and English: How does the model perform on other languages it has never seen in training? Does additional language model training in the target language improve performance?
    \item Combining the pretraining methods TSDAE and SimCSE: Both are unsupervised, but training a model using both methods either simultaneously or consecutively is possible and might improve performance.
    \item Splitting the model at some point during the training stages into an encoder model for report texts and an encoder model for requirement texts. Having dedicated models for both might increase matching performance due to further domain specialization.
\end{itemize}

We look forward to deploying the models developed during the course of this study in existing smart auditing software and expect a resulting optimization of machine learning workflows.

\section{Acknowledgements}
In parts, this research has been funded by the Federal Ministry of Education and Research of Germany and the state of North-Rhine Westphalia as part of the Lamarr-Institute for Machine Learning and Artificial Intelligence, LAMARR22B.

\nocite{*}
\bibliographystyle{ieeetr}
\bibliography{refs}

\begin{thebibliography}{10}

\bibitem{ali}
R.~Sifa, A.~Ladi, {\em et~al.}, ``Towards automated auditing with machine
  learning,'' in {\em Proceedings of the ACM Symposium on Document Engineering
  2019}, DocEng '19, (New York, NY, USA), Association for Computing Machinery,
  2019.

\bibitem{ramamurthy2021alibert}
R.~Ramamurthy, M.~Pielka, {\em et~al.}, ``Alibert: Improved automated list
  inspection (ali) with bert,'' in {\em Proceedings of the 21st ACM Symposium
  on Document Engineering}, DocEng '21, (New York, NY, USA), Association for
  Computing Machinery, 2021.

\bibitem{devlin2018bert}
J.~Devlin, M.-W. Chang, {\em et~al.}, ``{BERT}: Pre-training of deep
  bidirectional transformers for language understanding,'' in {\em Proc.
  NAACL-HLT}, 2019.

\bibitem{kpibert}
L.~Hillebrand, T.~Deußer, {\em et~al.}, ``{KPI-BERT}: A joint named entity
  recognition and relation extraction model for financial reports,'' in {\em
  Proc. ICPR}, 2022.

\bibitem{cao2018towards}
Y.~Cao, H.~Li, {\em et~al.}, ``Towards automatic numerical cross-checking:
  Extracting formulas from text,'' in {\em Proc. WWW}, 2018.

\bibitem{biesner2020leveraging}
D.~Biesner, R.~Ramamurthy, {\em et~al.}, ``Anonymization of german financial
  documents using neural network-based language models with contextual word
  representations,'' {\em Springer International Journal of Data Science and
  Analytics}, 2021.

\bibitem{tsai2017risk}
M.~Tsai and C.~Wang, ``On the risk prediction and analysis of soft information
  in finance reports,'' {\em European Journal of Operational Research}, 2017.

\bibitem{campbell2014information}
J.~Campbell, H.~Chen, {\em et~al.}, ``The information content of mandatory risk
  factor disclosures in corporate filings,'' {\em Review of Accounting
  Studies}, 2014.

\bibitem{reimers-2019-sentence-bert}
N.~Reimers and I.~Gurevych, ``Sentence-bert: Sentence embeddings using siamese
  bert-networks,'' in {\em Proceedings of the 2019 Conference on Empirical
  Methods in Natural Language Processing}, Association for Computational
  Linguistics, 11 2019.

\bibitem{transformer}
A.~Vaswani, N.~Shazeer, {\em et~al.}, ``Attention is all you need,'' 2017.

\bibitem{simcse}
T.~Gao {\em et~al.}, ``Simcse: Simple contrastive learning of sentence
  embeddings,'' 2021.

\bibitem{tsdae}
K.~Wang {\em et~al.}, ``Tsdae: Using transformer-based sequential denoising
  auto-encoder for unsupervised sentence embedding learning,'' 2021.

\bibitem{reimers-2020-multilingual-sentence-bert}
N.~Reimers and I.~Gurevych, ``Making monolingual sentence embeddings
  multilingual using knowledge distillation,'' in {\em Proceedings of the 2020
  Conference on Empirical Methods in Natural Language Processing}, Association
  for Computational Linguistics, 11 2020.

\bibitem{generating_financial_reports_from_tabular_data}
C.~L. Chapman, L.~Hillebrand, {\em et~al.}, ``Towards generating financial
  reports from tabular data using transformers,'' in {\em Machine Learning and
  Knowledge Extraction}, (Cham), pp.~221--232, Springer International
  Publishing, 2022.

\end{thebibliography}

\end{document}